\documentclass{article}

\usepackage{arxiv}

\usepackage[utf8]{inputenc} 
\usepackage[T1]{fontenc}    
\usepackage{hyperref}       
\usepackage{url}            
\usepackage{booktabs}       
\usepackage{amsfonts}       
\usepackage{nicefrac}       
\usepackage{microtype}      
\usepackage{lipsum}		
\usepackage{graphicx}
\usepackage{natbib}
\usepackage{doi}

\usepackage{amsmath}

\usepackage{chngcntr}
\usepackage[font=small,labelfont=bf]{caption} 

\title{Inter-model Interpretability: Self-supervised Models as a Case Study \thanks{The paper is under consideration at Computer Vision and Image Understanding Journal.}}

\author{{Ahmad Mustapha} \\
	Department of Electrical and Computer Engineering\\
	American University of Beirut\\
	Bliss Street, Beirut \\
	\texttt{amm90@mail.aub.edu}\\
	\And
{Wael Khreich}\thanks{Corresponding Author}\\
	Olayan School of Business\\
	American University of Beirut\\
	Bliss Street, Beirut \\
	\texttt{wk47@aub.edu.lb}\\
	\AND
	{Wes Masri} \\
	Department of Electrical and Computer Engineering\\
	American University of Beirut\\
	Bliss Street, Beirut \\
	\texttt{wm13@aub.edu.lb}\\
}

\renewcommand{\shorttitle}{\textit{arXiv} Template}

\hypersetup{
pdftitle={A template for the arxiv style},
pdfsubject={q-bio.NC, q-bio.QM},
pdfauthor={David S.~Hippocampus, Elias D.~Striatum},
pdfkeywords={First keyword, Second keyword, More},
}

\begin{document}
\maketitle

\begin{abstract}
	Since early machine learning models, metrics such as accuracy and precision have been the de facto way to evaluate and compare trained models. However, a single metric number doesn't fully capture the similarities and differences between models, especially in the computer vision domain. A model with high accuracy on a certain dataset might provide a lower accuracy on another dataset,  without any further insights. To address this problem we build on a recent interpretability technique called Dissect to introduce \textit{inter-model interpretability}, which determines how models relate or complement each other based on the visual concepts they have learned (such as objects and materials). Towards this goal, we project 13 top-performing self-supervised models into a Learned Concepts Embedding (LCE) space that reveals proximities among models from the perspective of learned concepts. We further crossed this information with the performance of these models on four computer vision tasks and 15 datasets. The experiment allowed us to categorize the models into three categories and revealed for the first time the type of visual concepts different tasks requires. This is a step forward for designing cross-task learning algorithms.
\end{abstract}


\section{Introduction}
\label{introduction}
Deep learning at its core is about automatic feature representation. Despite this, the current established approach for evaluating Deep Learning algorithms is to measure how they perform on certain tasks based on metrics such as accuracy or precision. While such metrics are useful they don't really tell how models relate from feature representation perspective. For example, two models that perform similarly from an accuracy perspective might have learned different complementary concepts. Confusion matrices on the other hand can give us more insights but they are limited to the training labels - bearing in mind that models can learn concepts associated with the labels without being directly trained on them (\cite{bau_understanding_2020}). This particularly not useful when comparing training algorithms rather than models. As in the latter case, we need to understand how the algorithm affect the feature representation learning process itself. Moreover, it goes without saying that in unsupervised learning setting there are no labels to compute a confusion matrix over.     

Evaluating models by the visual concepts they have learned complements the metrics approach and give us deeper insights. It allows us to match learned concepts to performance on different datasets and task. For instance, recently it has been shown that increased performance on ImageNet doesn't add any value to performance over a medical set when transferred. This means that the medical dataset requires learned concepts of different characteristics \citep{ke_chextransfer_2021}. 

\begin{figure}
  \includegraphics[width=1\textwidth]{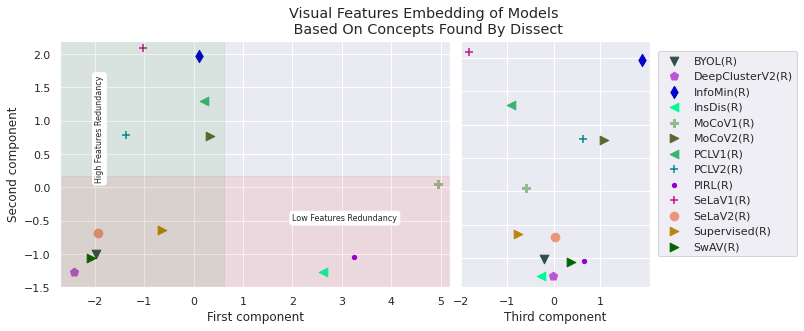}
\caption{The Learned Concepts Embedding (LCE) space of the 13 top performing self-supervised models based on their Dissect profile. Each dot represents a model. Three main groups emerge in the space.}
\label{fig:vfe}
\end{figure}

To address this problem we hereby introduce the notion of \textit{inter-model interpretability}. While traditional interpretability is concerned with: a) how a model makes a given decision, b) what the learned concepts are, or c) what fraction of the input drives the decision \citep{zhang_visual_2018}, Inter-model interpretability is concerned with: a) how a group of models relate to each other, b) whether the models learned similar concepts or different ones. Such analyses go beyond comparing models using solely evaluation metrics.

To develop this kind of interpretability we build upon a recent approach called Dissect \citep{bau_understanding_2020}. Dissect scalably inspects each unit (neuron) in a given network to understand the visual concept it has learned, then assigns the learned concept to the unit. Example concepts include, objects like horses and cars or materials such as fur and skin. It is possible to aggregate all the concepts learned by units in the final layer to come up with a learned concepts profile. We leverage such information by projecting models into a Learned Concepts Embedding (LCE) that emphasizes the proximities between the models from the perspective of what concepts they have learned.

As a case study, we projected the 13 top-performing self-supervised deep learning models into the LCE embedding\footnote{Data and code are available on GitHub repository: github.com/AhmadM-DL/Intermodel-Interpretation}. Self-supervised pretrained models are of interest because we can't directly compute an accuracy metric against the pretraining task as there is no ground truth available. The current approach is to freeze the model weights and evaluate the model against another task in a supervised learning setting. This approach is time-consuming and sensitive to the quality of the supervised learning procedure \citep{kolesnikov_revisiting_2019}. It is more insightful to complement the approach by considering directly the learned visual concepts. Our contributions can be summarised as follows:

\begin{itemize}
    \item We introduced the notion of inter-model interpretability and realized it by computing an embedding of the top-performing self-supervised models. As a result, we categorized the 13 top-performing models into three categories based on the visual concepts they learned in the LCE space. The embedding reveals the proximities of the models from the learned concepts perspective. We further explained each axis in the embedding in terms of a combination of different visual concept categories (objects, parts, materials, and colors). This allows researchers in the field to have an intuition about what type of visual concepts any new model retains based solely on its location in the embedding.

    \item By analyzing the correlation between the embedding axes and performance in several downstream tasks spanning 15 datasets, we found that some training algorithms are best suited for particular tasks based on the quantity and quality of the concepts they tend to construct in models. For example, we found that better localization in object detection tasks requires non-abstract shape detectors.

    \item Finally we investigated whether models that happen to be distant inside the embedding hold complementary visual concepts that would lead to better accuracy when combined. We found that this doesn't hold across tasks and datasets. This calls for an interpretation technique that is more representative and rich than Dissect. 
\end{itemize}

The following section provides background information on Dissect, self-supervised learning and deep learning tasks and reviews related literature. Section \ref{experimental setup} describes the experimental protocol including the models and datasets used. Section \ref{learned concepts} presents our proposed technique called the learned concepts embedding. The results are presented and analyzed in Section \ref{results}, followed by the conclusions and future work in Section \ref{conclude}.

\section{Background and Literature Review}
\label{background}

In this section, we introduce prerequisite concepts required for our proposal. We first present Dissect and describe its functionalities. Next, we provide an overview of the self-supervised workflow. Finally, we present a review of the relevant literature. 

\subsection{Dissect}
\label{dissect}
Dissect \citep{bau_understanding_2020} is a model interpretation technique that directly inspects the semantic knowledge a certain filter (unit) in a given network holds. To understand what activates a unit, a current practice in the literature is to feed the network a set of random images and manually inspect the images that highly activated it. This qualitative approach is useful but not scalable as it needs human interaction. The approach presented by Dissect scales this out by using a pretrained image segmentation model to replace human inspection.

\begin{figure}[h]
\begin{center}
       \includegraphics[width=0.7\linewidth]{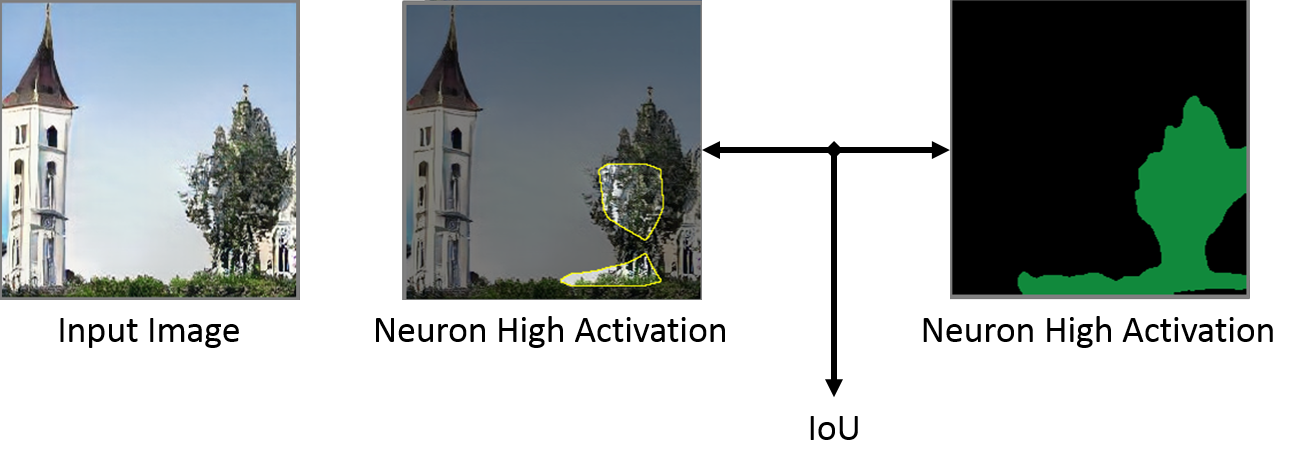}
\end{center}
   \caption{Dissect Basic Building Block.}
\label{fig:dissect}
\end{figure}

Consider the input image \citep{bau_understanding_2020} in Figure \ref{fig:dissect} and the corresponding up-scaled activation of a certain neuron on this image. It is evident that the neuron is looking for tree-like structures. But how would Dissect know that? It uses a pretrained image segmentor that segments the input image into different concepts (in Figure \ref{fig:dissect} it is segmenting the tree concept). It then computes the intersection over union (IoU) ratio between the neuron activation and the segmentor segmentation. To avoid chance scenarios it applies the same procedure to a large number of images in a validation dataset. If the IoU ratio is high over many images then this neuron is probably looking for the corresponding concept. Dissect computes the IoU ratio against 1,825 concepts available by the segmentor and assigns the concept with the highest ratio to the neuron. If the highest ratio is under a certain value, an IoU threshold hyperparameter, the neuron is reported to not belong to any concept (Check Appendix \ref{app:top} to see concepts that achieved top IoU for each considered model with images showing activations). 

\begin{figure}
\begin{center}
      \includegraphics[width=0.7\linewidth]{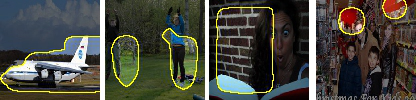}
\end{center}
  \caption{Example of units activating for different concept categories from left to right: airplane (object), leg (object part), brick (material), and red (color).}
\label{fig:example}
\end{figure}

Our concern in this paper, is that given a certain layer in a certain pretrained network, Dissect will output all the concepts that it captured in this layer along with the number of units that corresponds to each concept. The Dissect version  we used in this paper captures concepts of 4 different categories (subsequent versions of Dissect captures up to 6 categories): (1) Objects such as horse and person. (2) Object parts such as mountain-top and person-left. (3) Materials such as fur and skin. (4) Colors. Examples of units activating for such concepts can be seen in Figure \ref{fig:example}.

We will assign the term "Dissect Profile" for the result of applying Dissect on a certain layer in a given model. The result constitutes the list of detected concepts and the number of neurons assigned for each concept. We will also assign the term "Dissect Abstracted Profile" to the aggregation of Dissect profile by concept category. This profile is a vector of 4 values corresponding to the number of neurons assigned for each category: object, object part, material, color. When computing the abstracted profile we can consider all concepts or only unique concepts. In the first mode we count the number of neurons assigned to a category even if multiple neurons were assigned the same concept. In the second mode we count the number of unique concepts assigned to each category ignoring redundant concepts.

Finally it is important to note that Dissect's concept-assignment process is limited to an incomplete spectrum of concepts. For example despite the fact that some neurons captures pianos (check appendix \ref{app:example_activations}) no neuron was assigned to it. This limitation shouldn't shadow the fact that inter-model interpretability is a general methodology and not bound to using Dissect.

\subsection{Self-supervised Workflow}
\label{self-supervised}

Self-supervised deep learning (SSL) is a learning setting where the training labels are not available. To induce learning models are trained by solving a mock-up challenge.

Self-supervised representation deep learning (SSL) algorithms have gained attention by the community in recent years. They were introduced to overcome the limitations of supervised representation learning that requires a huge amount of expensive \textit{labeled} images and doesn't make use of the readily available unlabeled images. The ultimate goal of self-supervised learning algorithms is to be able to learn useful representation on par with the learned ones in the supervised setting or overcome them \citep{goyal_scaling_2019}.

In the SSL setting the models are trained by solving a mock-up challenge called a pretext task. The task is designed in a way that it could only be solved if the model learns useful abstract features. A wide set of SSL algorithms have been introduced with the main difference being the pretext task. Different tasks force the model to learn different levels of abstraction \citep{jing_self-supervised_2019}. Such a task can be coloring a grayscale image \citep{zhang_colorful_2016}, solving a jigsaw puzzle \citep{noroozi_representation_2017}, contrasting a set of images \citep{he_momentum_2020}, grouping images together \citep{caron_deep_2019}, and so on. Clustering-based and Contrastive-based approaches are, to the date of writing this paper, the best-performing tasks \citep{ericsson_how_2020}.

In order to evaluate the performance of an SSL algorithm, the process goes as follows: (1) The model is pretrained in an unsupervised fashion over the pretext task. (2) The learned features are used in a supervised setting over what is called a downstream task. The task usually entails training a linear classifier, however, in some cases it also entails object detection or segmentation. The best SSL algorithm is the one that outperforms the others on a wide range of downstream tasks.

\subsection{Deep Learning Downstream Tasks}
\label{downstream tasks}
For completeness, this section briefly describes the four different downstream tasks considered.

\textbf{Many Shot Classification}. A widely used learning setting in which a model learns to classify a set of inputs into two or more classes while having an abundance of labeled inputs. They are usually trained by minimizing the cross-entropy between predictions and ground truth.

\textbf{Few Shot Classification}. A learning setting in which a model learns to classify a set of inputs into two or more classes while having a very limited set of labeled inputs usually ranging between 5 to 50 for each class. It considers the concept that reliable algorithms should learn to perform well using a small amount of training data. Here we consider prototypical networks. The process entails two steps: (a) A pretrained model is used as a non-linear mapping of the training set and the query set into an embedding space. (b) A prototype is assigned for each class by computing the mean of its support set (training set for each class) in the embedding space. Prediction is done, given an embedded query point (an unlabeled input), by simply finding the nearest class prototype.

\textbf{Object Detection}. A learning setting in which a model learns to not only classify objects in images but also detect their bounding box. Different evaluation metrics exist each giving different priority for classification and localization. 

\textbf{Surface Normal Estimation}. A learning setting in which a model learns to assign the normal (a vector perpendicular to the surface at that point) for each pixel in an image. Given, for example, a 2-dimensional image of a table. The tabletop pixels will be assigned a normal perpendicular to the x-y plane.

\begin{figure}[h]
\begin{center}
      \includegraphics[width=0.44\linewidth]{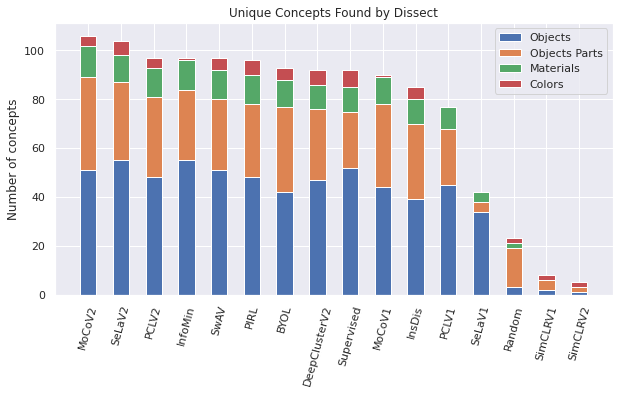}
       \includegraphics[width=0.44\linewidth]{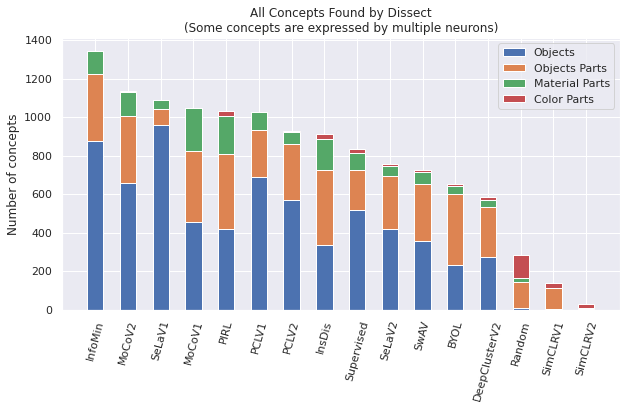}
\end{center}
  \caption{The number of unique/all concepts in each category found by Dissect for each model.}
\label{fig:dissect_profile}
\end{figure}

\subsection{Related Work}

Our work in this paper is related to both the interpretable and the self-supervised deep learning domains. Considering the latter, our work is positioned along with other work that tries to analyze the performance of self-supervised models. Goyal et al \citep{goyal_scaling_2019} studied how the performance of different self-supervised learning algorithms changes with scaling both the pretraining data size and the model's capacity. Similarly, Kolesnikov et al. \citep{kolesnikov_revisiting_2019} studied the effect of different factors involved in such algorithms as widening models, and linear evaluation protocol. Ericsson et al. \citep{ericsson_how_2020} conducted a comprehensive study using a benchmark of top-performing models along different downstream tasks.

With respect to interpretable deep learning, the domain started to form in response to concerns about the black box nature of neural networks and the lack of interpretability. The domain have grown significantly and diverse in recent years. The mainstream interpretability approaches are: (a) Feature visualization, which tries to generate examples that maximally activate part of the network let it be a unit, a channel, a layer, or class probability \citep{erhan_visualizing_2009, simonyan_deep_2014, zeiler_visualizing_2014}. (b) Feature attribution, which tries to explain which part of the image contributed mostly to the network decision \citep{selvaraju_grad-cam_2017, lundberg_unified_2017, sundararajan_axiomatic_2017}. Olah et al. \citep{olah_feature_2017} presented an interactive toolbox. Zhang et al. \citep{zhang_visual_2018} compiled a  survey on visual interpretation of CNNs. 

Other researchers went on to explain predictions by case, i.e. by proposing a case that is believed by the targeted neural network to be closest to a query case \citep{chen_this_2019, bien_prototype_2011}. Or by generating textual explanation \citep{hendricks_generating_2016, karpathy_deep_2017}. A comprehensive study is presented by Fan et al. \citep{fan_interpretability_2021}.

As far as we know we are the first to consider inter-model interpretability. We used the recent interpretability technique Dissect \citep{bau_understanding_2020} to understand how top performing self-supervised models are similar to each other. We also linked the availability of certain concepts categories such as objects and materials to performance on different downstream tasks and datasets.

\section{Experimental Setup}
\label{experimental setup}

In our experiment, we applied Dissect on 13 models pretrained using top-performing self-supervised algorithms. All the models use ResNet50 \citep{he_deep_2015} as a backbone and are pretrained over ImageNet. We also considered a ResNet model trained in supervised fashion over ImageNet for comparison. And finally we considered a ResNet model with random weights as a baseline. We used Dissect authors code as is, including the validation Places dataset \citep{zhou_learning_2014}, the Unified Perceptual Parsing image segmentation network \citep{xiao_unified_2018}, and the IoU threshold as 0.04. We did; however, try different thresholds and found that the proposed threshold strikes a balance. Increasing the threshold will decrease severely the number of neurons with assigned concepts. Decreasing the threshold will lead to assigning a lot of neurons that sparsely fires for the assigned concept. We considered in our experiment only the final layer in ResNet50 which constitutes 2048 neurons.

The models used in our experiment are InsDis, MocoV1/V22, PIRL, SimCLRV1/V2, InfoMin, and BYOL as contrastive-based SSL. PCLV1/V2, SeLaV1/V2, DeepClusterV2, SwAV as clustering-based SSL.

To enrich our data we crossed the generated Dissect profiles of the models with their performance on a set of different downstream tasks and datasets from \citep{ericsson_how_2020}. The Datasets used are: FGVC Aircraft,  Stanford Cars, Oxford 102 Flowers, Food-101, and Oxford-IIIT Pets as many-shots fine-grained object classification. ImageNet, Caltech-101,  CIFAR-10, CIFAR-100, and Pascal VOC2007 as many-shots coarse-grained object classification. DTD as many-shot texture classification. And SUN397 as many-shot scene classification. CropDiseases, EuroSAT, ISIC2018 and ChestX as few-shots classification reporting on 5-10-50 shots for each. Pascal VOC2007  as detection reporting AP, 50AP, 75AP with and without fine-tuning. NYUv2 as dense prediction of surface normal estimation.

In this manuscript we don't aim to find the best performing self-supervised model on all available datasets and tasks, but we focus on the learned concepts as detailed in the following sections. Interested readers can check \citep{ericsson_how_2020} for more details about the training procedure, evaluation metrics, and models' performance on each task.

\section{Learned Concepts Embedding}
\label{learned concepts}
In order to compute the LCE, we first found the super-set of all concepts found by Dissect in all aforementioned models. We then represent each model by a vector where each dimension represents a concept in the super-set. The dimensions values represent the number of units Dissect matched to the concept in the model's last layer. We call this vector for each model a\textit{ Dissect Profile}. We finally normalized the values with the total number of concepts found for each category. The super-set included 144 concepts distributed as following 67 objects, 54 object parts, 16 materials, and 8 colors. Formally, consider that we have a set of models $M=(m_1, m_2, ..,)$, a super-set of concepts $C=(c_1, c_2, ..)$. Let $D(m_i, c_j)$ be the number of units found by Dissect in model $m_i$ that is matched to concept $c_j$. Let $(t_o, t_p, t_m, t_c)$ be the total number of objects, object parts, materials, and colors respectively that compose the super-set $C$. Let $cat(c_j)$ be a mapping of the concept $c_j$ to its corresponding category. Then the LCE will be computed over the following matrix:

$$
\begin{bmatrix}
\frac{D(m_1, c_1)}{t_{cat(c_1)}} & .. & .. & .. \\
.. & .. & .. & .. \\
.. & .. & \frac{D(m_i, c_j)}{t_{cat(c_j)}}  & ..\\
.. & .. & .. & .. \\
\end{bmatrix} 
$$

The LCE is computed by applying Principle Component Analysis (PCA) over the aforementioned matrix. We considered only the first three components because they explained 88\% of the total variance. The scatter plot of the embedding can be seen in Figure \ref{fig:vfe} along with the different models considered in our experiment. 

\section{Results and Discussion}
\label{results}

\subsection{What Each Self-supervised Model Have Leaned}
\label{dissectprofiles}

As aforementioned Dissect was applied on 13 Self-supervised models along with a supervised model and one with random weights. Out of the 2048 neurons in the last layer of ResNet50 some where assigned a visual concept in each model. The abstracted Dissect profile of the considered models can be seen in Figure \ref{fig:dissect_profile}.

The Random model has a simple profile composed of a handful of concepts. However, visually inspecting corresponding top activation for each neuron shows that they are random and don't hold any semantics. Those random activations tend to intersect with segments of frequently occurring concepts in the validation dataset such as sky, floor, and person. This points out a major weakness in Dissect algorithm. On the other hand, SimCLR models have yet a smaller profile than a random model and detect similar concepts. Given that SimCLR models have a very good performance, it seems that it encodes information in a different way than the other models. For the rest of the models, the number of unique concepts found doesn't vary a lot across the majority of models. MoCoV2 and SeLaV2 are leading. Insdis, PCLV1, and SeLaV1 are trailing.

The models differed more in the total number of concepts assigned to neurons. For instance, while the supervised model have been found to detect approximately 90 unique concepts similar to BYOL and DeepClusterV2, they where distributed over more than 800 neurons unlike the latter which where distributed to less than  700 neurons. This assignment redundancy is a Dissect artifact an it doesn't necessarily mean that all those neurons assigned to the same concept capture only the concept itself. We will further discuss redundancy later in the manuscript.

\begin{figure}[h]
\begin{center}
      \includegraphics[width=0.7\linewidth]{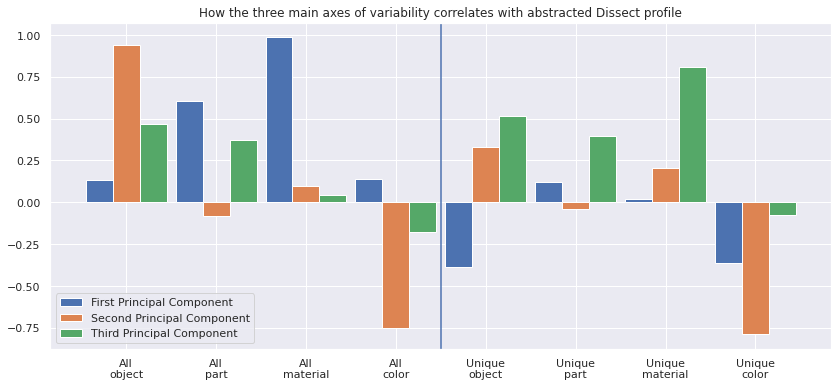}
\end{center}
  \caption{Correlation between abstracted Dissect profile and principal components.}
\label{fig:understand_pca}
\end{figure}

\subsection{Explaining The Embedding}
\label{explain embedding}

Hereby we present our effort to find the variance factor of each principal component.  In other words, we want to find how the number of all or unique concepts detected by Dissect effects the model's location in the embedding. For this we computed the correlation of each concept category with the the three principal component. The results are found in Figure \ref{fig:understand_pca}. 

The first component is highly correlated with the total number of materials and to the total number of object parts with a lesser degree. The same pattern occurs for the second component but it considers objects and color. On these axes, the number of unique concepts found by Dissect doesn't matter as much as how many units represent a given concept. This is expected as the majority of models tended to have similar unique concepts. Moreover, the second component has a negative correlation with color which means it also represent color information. The third component correlates mostly with the number of unique materials found. Unlike the first two components, it considers the number of unique concepts to a certain degree.

In summary, the embedding's first axis looks for materials and object parts and the second axis looks for objects and colors. The variance on both axes represents how many units are matched to the corresponding categories of concepts. The third axis looks for the number of unique concepts found with more emphasis on materials. As a result we now can have an intuition about what type of visual concepts a model pertain based on its location in the embedding.

\subsubsection{Reading The Embedding}
\label{read embedding}

The projection of the Dissect profiles for each model resulted in a space of three clusters as shown in Figure \ref{fig:vfe}. Note that we run KMeans with varying number of clusters on the Dissect profiles data to find the best number of clustering according to the elbow method and using inertia as quality metric. We found that choosing three clusters is the best (see Appendix \ref{app:elbow}). The first cluster (A) resides in the lower region of the two principal components. The second cluster (B) residing in the higher end of the first component and the lower end of the second component. (C) residing in the higher end of the second component and the lower end of the first component.

We know from the previous section that all models in all sections have similar unique concepts profile but differ in the number of neurons assigned. Models in (B) have more neurons assigned to materials, while models in (C) have more neurons assigned to objects.
This can be interpreted only if we considered the aforementioned limitation of Dissect which states that Dissect can only capture limited number of concepts. The low number of assigned neurons means that other un-assigned neurons are actually abstract representation of concepts that Dissect doesn't cover. This case corresponds to models in (A). On the other hand, the high number of neurons are assigned to same concept means that the neurons are different non-abstract representation that happen to fire for the same concept. Based on this models in (B) are rich in non-abstract pattern detectors as the first component is correlated with materials category, and the models in (C) are rich in non-abstract shape detectors as the second component is correlated with objects category.

\begin{figure}[h]
\begin{center}
      \includegraphics[width=0.7\linewidth]{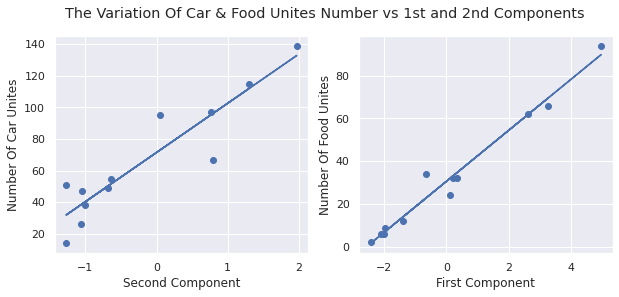}
\end{center}
  \caption{How the number of food and car units varies with the 1st \& 2nd components respectively.}
\label{fig:car_food}
\end{figure}

To emphasis on this more, consider the concept of car (object) and food (material). In Figure \ref{fig:car_food} we plot how the number of neurons matched to the food and car concepts varies with respect to the first and second components respectively. Some models have more neurons that detect cars because the model hasn't reached to a certain level of abstraction to have only one neuron to detect a car. In other words, units that detect wheels, smoothed edges, mirrors, or geometric shapes will end up being matched by Dissect to car concept for they mostly intersect with car concept. The same is true for the food concept on the second component.

From an algorithmic perspective Cluster (A) is dominated by clustering-based approaches mainly SeLaV2, SwAV, and DeepClusterV2. It also contains BYOL and Supervised. All these models learn by grouping instances despite the nature of grouping (clusters, labels, or augmentation) without using negative samples. The other approaches spread in clusters (B) and (C) are contrastive approaches that do use negative samples. Except for SeLaV1 which is clustering-based and precursor to SeLaV2. All the V2 models are derived from V1 models mainly by adding projection heads, more augmentations, and more training epochs. This operation has moved SeLa from cluster (C) to cluster (A), MoCo from cluster (B) to (C), and has moved PCL mostly on the third component. We can deduce that some algorithms allows the model to favor shape or pattern concepts detectors, while others strike a balance between favoring both shape and pattern detectors.

\subsection{What are the Concepts Required for each Task?}

\begin{figure}[h]
       \includegraphics[width=1\linewidth]{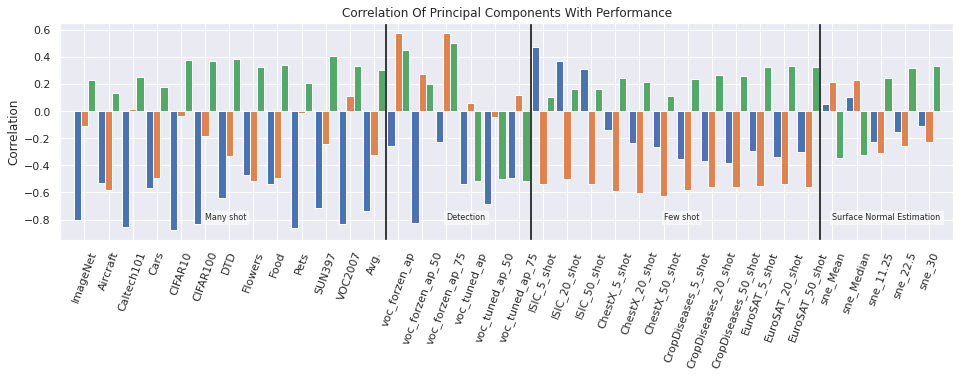}
  \caption{Correlation between embedding directions and performance.}
\label{fig:direction}
\end{figure}

We computed the correlation between the three principal components and the performance of different learning tasks. Does a higher value for a given component means better performance on some tasks. The goal of this experiment is to understand the importance of the embedding different components  on different types of tasks. The results are shown in Figure \ref{fig:direction}. 

For Many Shot classification tasks the performance has a negative correlation with the first component. For such tasks, it is important to have abstract detectors or non-abstract shape detectors. To the contrary, when the task is fine-grained such as in Aircraft and Cars, performance becomes more correlated with the second component. Which requires balanced profile of abstract detectors.

Unlike many shot classification, few-shot classification is more negatively correlated with the second axis. For such tasks, it is important to have abstract detectors or non-abstract pattern detectors. ISIC dataset in particular has a weak positive correlation with the first axis. This means it requires non-abstract pattern detectors. This is expected for the ISIC dataset, skin lesion images are visually very different from natural images in ImageNet. 

For detection, however, an interesting insight emerges. When we are training with an emphasis on classification more than on localization i.e. using AP50 metric, performance is correlated negatively with the first axis as it requires abstract detectors. However, when we give more importance to localization using stricter metrics such as AP and AP75 performance become positively correlated with the second and third axis. It require non-abstract shape detectors with diversity of detectors. To make this concrete, a horse detector won't help the model to localize things but some shape detectors well.When we consider fine-tuning this become less important.

Performance on surface normal estimation has a weak correlation with all axes. This task requires certain features that are not captured by Dissect. Possible features such as edges or depth of field.

As a result we can now, after projecting models into the embedding space, decide which models are useful or not for a specific tasks based solely on their location in the embedding (see Appendix \ref{app:detailed_performance_embedding}).

To generalize on this analysis, if we need to train a model in an unsupervised setting to be used as an object detector backbone we should look for algorithms that generates many non-abstract shape detectors. Out of the considered algorithms we have MoCoV2 and Info-Min to achieve this combination particularly given that have the largest number of assigned neurons in Dissect profile (check Figure \ref{fig:dissect_profile}). Same reasoning applies for other tasks and dataset characteristics. 

\subsection{How Visual Detectors Complement Each Other}
\label{complement}

Does distant models in the LCE complement each other? To answer this question we performed the following experiment. We first measured the complementary information between models. For each couple of models, we build a weighted soft voting ensemble. The models' probability predictions are weighted based on the performance of individual models. Let $a$, and $b$ with $a>b$ be the performance of models $m_1$ and $m_2$ respectively. Then the prediction of models ensemble: $$E_{m_1,m_2}(X) = (0.5+|a-b|)m_1(x) + (0.5 -|a-b|)m_2(x)$$
We choose this simple ensemble method for (1) The low performing model doesn't dominate the better performing one (2) We need any good enough ensemble to generate a complementarity signal. We performed our analysis on three different datasets (DTD, Aircraft, and Caltech101). 

\begin{figure}[h]
\begin{center}
           \includegraphics[width=1\linewidth]{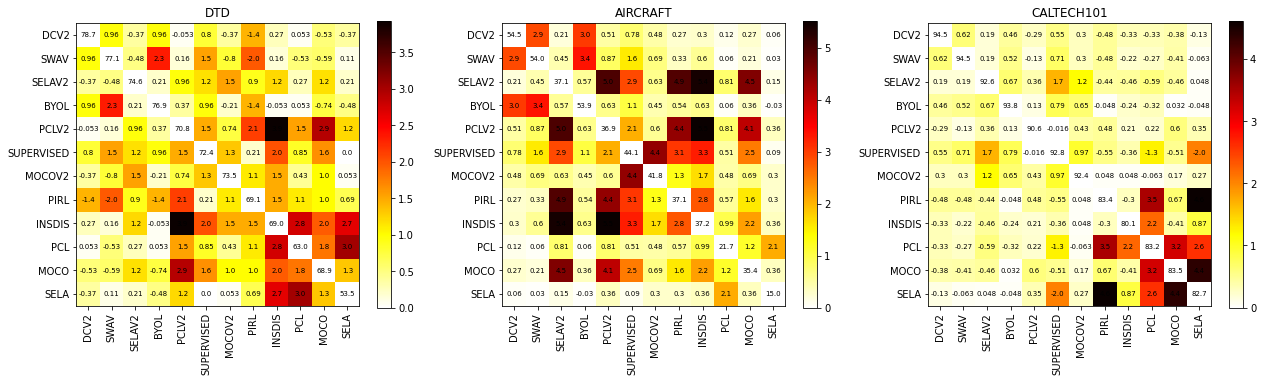}
\includegraphics[width=1\linewidth]{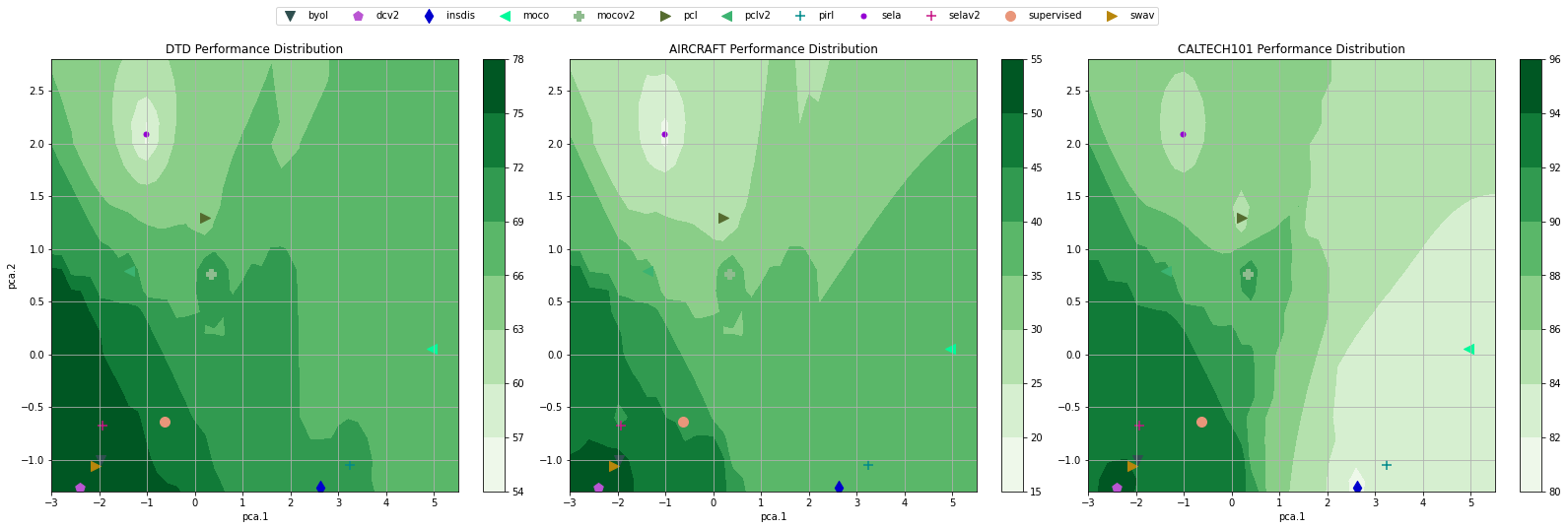}
\end{center}
  \caption{Ensemble Gain/Loss Results (Top) And Performance Distribution (Bottom) For Different Datasets.}
\label{fig:ensemble}
\end{figure}

The result of the ensembles are presented as a heatmap in Figure \ref{fig:ensemble}. The heatmap diagonal reports the classification accuracy of each model on the dataset. The other cells report the performance gain/loss after combining each corresponding pair of models. Given two models pairs (A, B) the reported values is $Perf_{Ensemble(A,B)} - Max(Perf_A, Perf_B)$. The darker the color, the  better the ensemble.

We then visually inspected how ensemble performance is affected by models placement in the LCE. Alongside the heatmaps we plot the performance distribution of the models over the embedding space. The performance at each point in the embedding is interpolated based on the five nearest neighbors performance.

\subsubsection{DTD}
The DTD dataset is a fine-grained texture classification dataset. The most notable gain in performance is associated with models in group (B): InsDis, MOCOV1 along with PCLV1 and PCLV2. These models achieved the highest performance improvements when combined with each other. One interesting insight is that the Supervised model was also able to achieve an improvement in performance when combined with some of these models. Swav and DeepClusterV2 are top performing models and they only achieved an improved performance when combined with each other. By looking at the diagonal of the heatmap (Figure \ref{fig:ensemble}), we can see that the most benefiting models have low performance except for PCLV2.

\subsubsection{Aircraft}
Aircraft is also a fine-grained classification dataset but unlike DTD it requires more sophisticated features. The combinations that achieved the highest improvements in performance included SeLaV2 and PCLV2. Furthermore, the combination of Supervised with InsDis, PIRL, and MoCoV1 increased the ensembles performance. For the remaining models, they showed an improvement in performance when combined with models from their groups. This reveals the fact that SeLaV2, PCLV2, and Supervised all favor a high level of feature abstraction  in their training algorithms. On the contrary, SwAV, BYOL, and DeepClusterV2 maintain a balance between high and low level of feature abstraction at the end of their training process.

\subsubsection{Caltech101}
The Caltech101 dataset is a coarse-grained natural image for classification tasks, it is similar to the ImageNet dataset which has been used to train all models considered in this paper. 

Except for PCLV1 and SeLAV1, which have extreme values regarding the third component, ensembles are not able to improve the accuracy over single models. Only the aforementioned models which belong to group (C) achieved a considerable level of accuracy improvement after being combined with models from group (B): InsDis, PIRL, and MoCoV1. This might be due to the fact that all the models (considered in our work) have been trained over ImageNet, which have similar characteristics to Caltech101, leading to a performance saturation where ensembles are not adding more information to each model feature set.

From the above experiments, we can conclude that the characteristics of the dataset play an important role in providing complementary information for the combined models. This means understanding the targeted visual concepts that are included in different datasets is a key for extracting complementary feature profiles. In general, combining accurate and diverse models (which learn various concepts and potentially commit different types of errors) could lead to improved ensemble performance. In our case, the SWAV, BYOL, and DeepClusterV2 models achieved the best performance on several datasets with different profiles due to their common underlying clustering algorithm. For a conclusion the experiment show that distant models in the LCE don't necessarily complements each other.

\section{Conclusion and Future Work}
\label{conclude}

In this paper, we introduce the notion of inter-model interpretability to study the similarities and differences among various deep learning models based on concepts learned during training. We investigated the main differences between 13 self-supervised models by projecting them into a learned concept embedding (LCE) space. We demonstrated that each axis in the embedding space comprises a different category of concepts and discovered that these models can be divided into three main groups.  By intersecting the information provided by the LCE with models' performance on 15 datasets over four different downstream tasks, we deduced what combination of concepts each task requires. As such, Inter-model interpretability is interesting and valuable for designing models for specific tasks.

Even though Dissect has captured important axes of differences among the studied models, it focuses on a limited number of visual concepts, which doesn't cover the entire concept spectrum learned by the deep learning model. An interesting future work direction would be to increase Dissect coverage by including more visual concepts such as blur, depth, etc. On the other hand, it is worth investigating inter-model interpretation approaches that do not depend on Dissect profiles, such as comparing the activation between two models directly.

Another interesting future goal is to apply Dissect to a larger pool of models with different training procedures, architectures, training data, and training tasks. This will enhance the LCE quality and give more insights into the interaction between performance and learned concepts space. Furthermore, it would be useful to study the effect of different hyperparameters such as dropout, normalization, capacity, and activation functions on models' Dissect profiles in the LCE space.


\bibliographystyle{unsrtnat}
\bibliography{references}

\newpage

\section*{Appendix}
\renewcommand{\thesection}{\Alph{section}}
\setcounter{section}{0}
\counterwithin{figure}{section}

\section{Top Matching Units} \label{app:top}
In the below Figure \ref{fig:top_matches} we report the top 5 matching units along with the concept they were matched to and the IoU value using Dissect for each model. Random and SimCLR models units tend to be matched coincidentally.

\begin{center}
\begin{minipage}{\linewidth}
\noindent\makebox[\textwidth]{%
\includegraphics[scale=0.3]{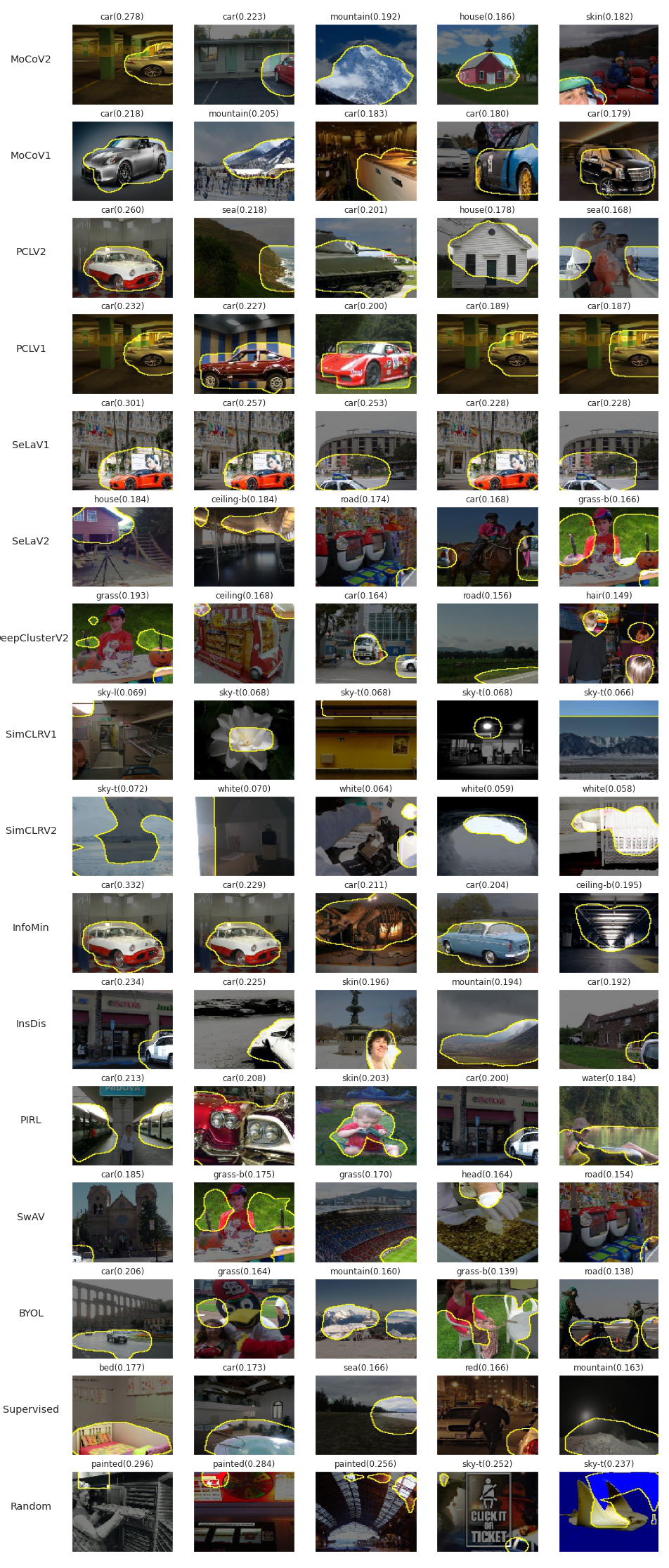}}
\captionof{figure}{Top 5 Dissect's matching units for each model}
\end{minipage}
\label{fig:top_matches}
\end{center}

\section{Exemplary Activations} \label{app:example_activations}
The first row of images in the below Figure \ref{fig:example_activations} corresponds to a unit in the supervised model. The unit is matched to the swimming pool concept. The second row corresponds to a unit in PCLV1 model. The unit is matched to no concept by Dissect.

\begin{center}
\begin{minipage}{1\linewidth}
  \includegraphics[width=1\linewidth]{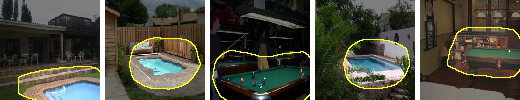}
  \includegraphics[width=1\linewidth]{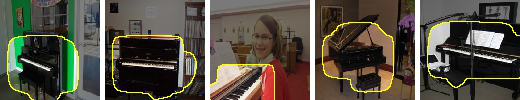}
  \captionof{figure}{Dissect Limitations}
\end{minipage}
  \label{fig:example_activations}
\end{center}


    
    


\section{Principal Components Coefficients} \label{app:coefficients}

After applying Principal Component Analysis on normalized vectors of Dissect concepts we consider the top three components which represent ~87\% of data variance. The below Figure \ref{fig:pca_detail} represents the detailed concept composition of each component. As you can see the first component has the materials fur and skin as major concepts along with other object parts. While the second component has more emphasis on objects. The best performing models tend to minimize both components balancing between different abstractions.

\begin{center}
\begin{minipage}{1\linewidth}
  \noindent\makebox[\textwidth]{%
  \includegraphics[width=1.1\linewidth]{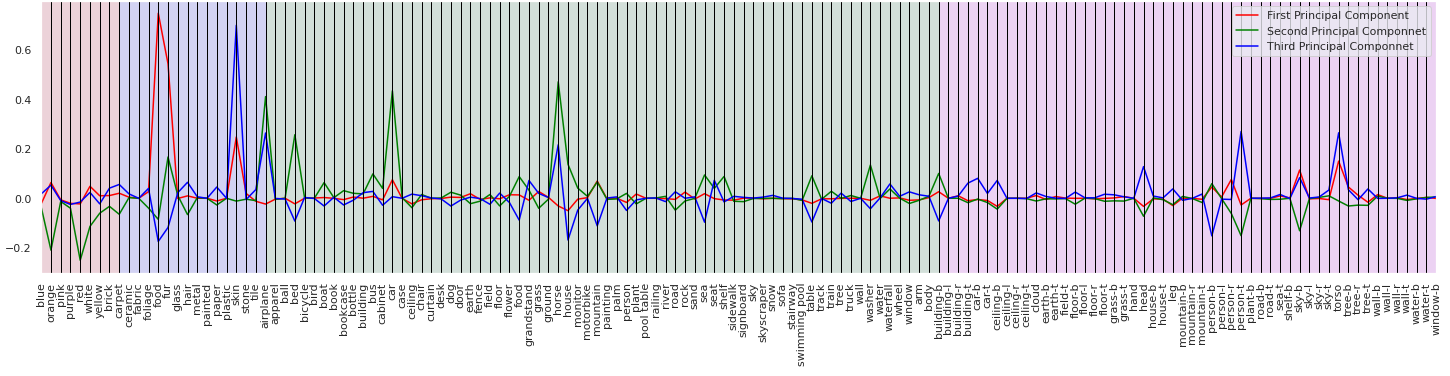}}
  \captionof{figure}{The Detailed Composition Of Dissect Main Axes Of Variance}
\end{minipage}
  \label{fig:pca_detail}
\end{center}

\vfill

\section{Clustering Elbow Method}\label{app:elbow}

In Figure \ref{fig:elbow} we report the application of the elbow method over dissect profiles to decide the best number of clusters to represent the embedding. According to the method the best fitting number of clusters is three.

\begin{center}
\begin{minipage}{1\linewidth}
  \noindent\makebox[\textwidth]{%
  \includegraphics[width=0.6\linewidth]{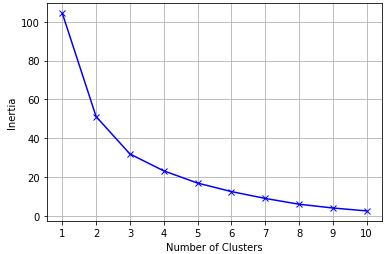}}
  \captionof{figure}{The Elbow Method Used To Decide Number Of Clusters}
\end{minipage}
  \label{fig:elbow}
\end{center}

\section{Performance Vs Embedding Position}\label{app:detailed_performance_embedding}

In Figure \ref{fig:perf_embed} we report the relation between the position in the LCE and performance on different dataset-task pairs (Performance interpolation was done using a KNN regressor with k=5). It is evident that the 
LCE do reflect partial feature information that dictates performance in different tasks. For instance Many Shot Classification tasks in particular those who have images similar to ImageNet such as CIFAR10, CIFAR100, and VOC2007 tends to achieve better performance when the model is in the button left corner of the embedding. The pattern is different for the detection task if we give more importance for localization (AP, and AP75) where the central models tends to perform better. This means that there is a fundamental difference between features that are useful for classification and those which are useful for localization. Same applies to few shot learning tasks on ISIC dataset in particular where models in the button right corner best performs.

\vfill

\begin{center}
\begin{minipage}{\linewidth}
  \noindent\makebox[\textwidth]{%
  \includegraphics[width=0.8\linewidth]{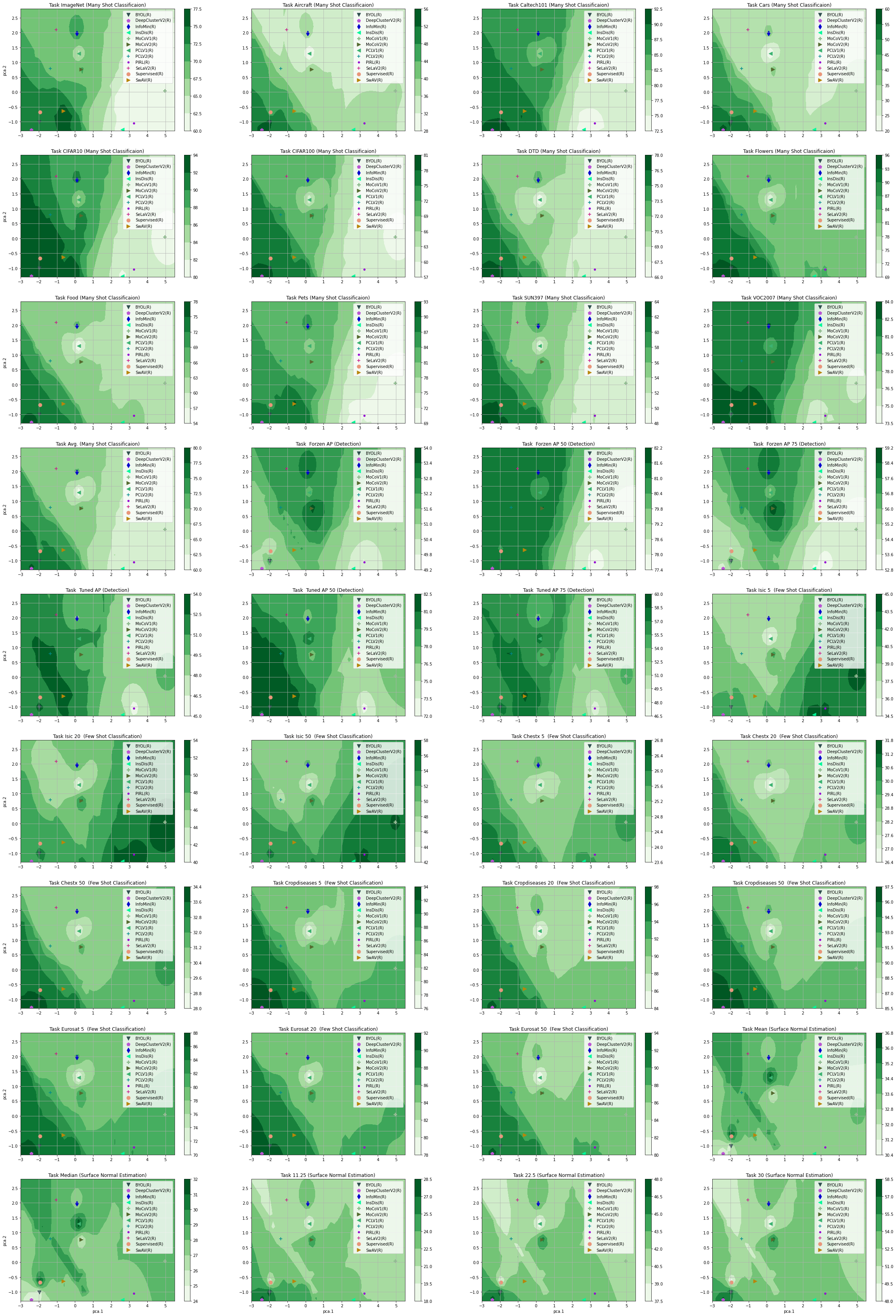}}
  \captionof{figure}{The Relation Between Placement in LCE and Performance on Several Tasks}
\end{minipage}
  \label{fig:perf_embed}
\end{center}

\end{document}